\documentclass[runningheads]{llncs}

\usepackage{eccv}

\usepackage{amsmath,amsfonts,bm}

\def\eqref#1{equation~\ref{#1}}

\def\1{\bm{1}}

\DeclareMathAlphabet{\mathsfit}{\encodingdefault}{\sfdefault}{m}{sl}
\SetMathAlphabet{\mathsfit}{bold}{\encodingdefault}{\sfdefault}{bx}{n}

\usepackage{amsmath}
\usepackage{amsfonts}
\usepackage{amssymb}
\usepackage{wrapfig}
\usepackage{subcaption}
\usepackage{multirow}
 \usepackage{mathtools} 

\usepackage{verbatim}

\usepackage{anyfontsize}

\usepackage{microtype}
\usepackage{graphicx}
\usepackage{booktabs} 
\usepackage{multirow}
\usepackage{amsmath,amssymb}
\usepackage{booktabs}
\usepackage{caption,subcaption}

\usepackage{xcolor}
\definecolor{mygreen}{HTML}{3cb44b}
\definecolor{skyblue}{HTML}{beffff}
\definecolor{lightgreen}{HTML}{90ee90}

\usepackage{color, colortbl}

\definecolor{emerald}{rgb}{0.31, 0.78, 0.37}

\usepackage{tcolorbox}
\usepackage{enumitem}
\setitemize{itemsep=10pt,topsep=0pt,parsep=0pt,partopsep=0pt}
\usepackage{colortbl}

\usepackage{xcolor}
\definecolor{mygreen}{HTML}{3cb44b}
\colorlet{myyellow}{green!10!orange!90!}
\makeatletter

\usepackage{tikz}
\usetikzlibrary{arrows,shapes,snakes,automata,backgrounds,fit,petri}
\usepackage{adjustbox}

\newcommand{\RN}[1]{%
	\textup{\lowercase\expandafter{\it \romannumeral#1}}%
}
\usepackage{tabu}

\newcommand{\beq}{\vspace{0mm}\begin{equation}}
\newcommand{\eeq}{\vspace{0mm}\end{equation}}
\newcommand{\beqs}{\vspace{0mm}\begin{eqnarray}}
\newcommand{\eeqs}{\vspace{0mm}\end{eqnarray}}
\newcommand{\barr}{\begin{array}}
\newcommand{\earr}{\end{array}}

\usepackage{color, colortbl}
\definecolor{Gray}{gray}{0.93}

\usepackage{lipsum}

\usepackage{pifont}
\newcommand{\cmark}{\ding{51}}%
\newcommand{\xmark}{\ding{55}}%

\usepackage{makecell}

\usepackage{xcolor,amsmath}
\usepackage[linesnumbered,ruled,vlined]{algorithm2e}
\DontPrintSemicolon

\usepackage{xcolor}
\definecolor{mygreen}{HTML}{3cb44b}

\SetKwComment{Comment}{\color{green!50!black}\# }{}

\SetKwProg{Function}{def}{:}{}

\SetKwProg{For}{for}{:}{}
\SetKwProg{If}{if}{:}{}

\usepackage{eccvabbrv}
\usepackage{graphicx}
\usepackage{booktabs}
\usepackage[accsupp]{axessibility}  
\usepackage{hyperref}
\usepackage{orcidlink}
\usepackage[utf8]{inputenc}
\usepackage{array, caption, floatrow, makecell, booktabs}
\usepackage{wrapfig}
\usepackage{floatrow}
\usepackage{comment}
\usepackage{float}
\usepackage{wrapfig,lipsum,booktabs}
\usepackage{tabularx} 
\usepackage{tabu}
\usepackage[capitalize]{cleveref}
\crefname{section}{Sec.}{Secs.}
\Crefname{section}{Section}{Sections}
\Crefname{table}{Table}{Tables}
\crefname{table}{Tab.}{Tabs.}
\usepackage{wrapfig}
\usepackage{multirow,array}
\usepackage{xcolor}
\usepackage{color, colortbl}
\definecolor{Gray}{gray}{0.93}

\newlength\savewidth

\newcommand\extrafootertext[1]{%
    \bgroup
    \renewcommand\thefootnote{\fnsymbol{footnote}}%
    \renewcommand\thempfootnote{\fnsymbol{mpfootnote}}%
    \footnotetext[0]{#1}%
    \egroup
}

\usepackage{listings}
\usepackage{xcolor}

\usepackage{multirow}
\usepackage{siunitx} 
\usepackage{caption} 
\usepackage{pifont}

\makeatletter
\def\fps@table{tbp}
\def\fps@figure{tbp}
\makeatother

\usepackage[section]{placeins}

\begin{document}

\title{Multi-label Instance-level Generalised Visual Grounding in Agriculture}

\author{Mohammadreza Haghighat\inst{1,2}\thanks{Corresponding authors.}, Alzayat Saleh\inst{1,2}, Mostafa Rahimi Azghadi\inst{1,2}$^{\star}$}
\authorrunning{M. Haghighat et al.}
\institute{
    College of Science and Engineering, James Cook University, Townsville, QLD, Australia \\
    \and
    Centre for AI and Data Science Innovation, James Cook University, Townsville, QLD, Australia \\
    \email{reza.haghighat@my.jcu.edu.au, mostafa.rahimiazghadi@jcu.edu.au}
}



\maketitle

\vspace{-5mm}

\begin{figure*}[!ht]
    \centering
    \includegraphics[width=1\linewidth]{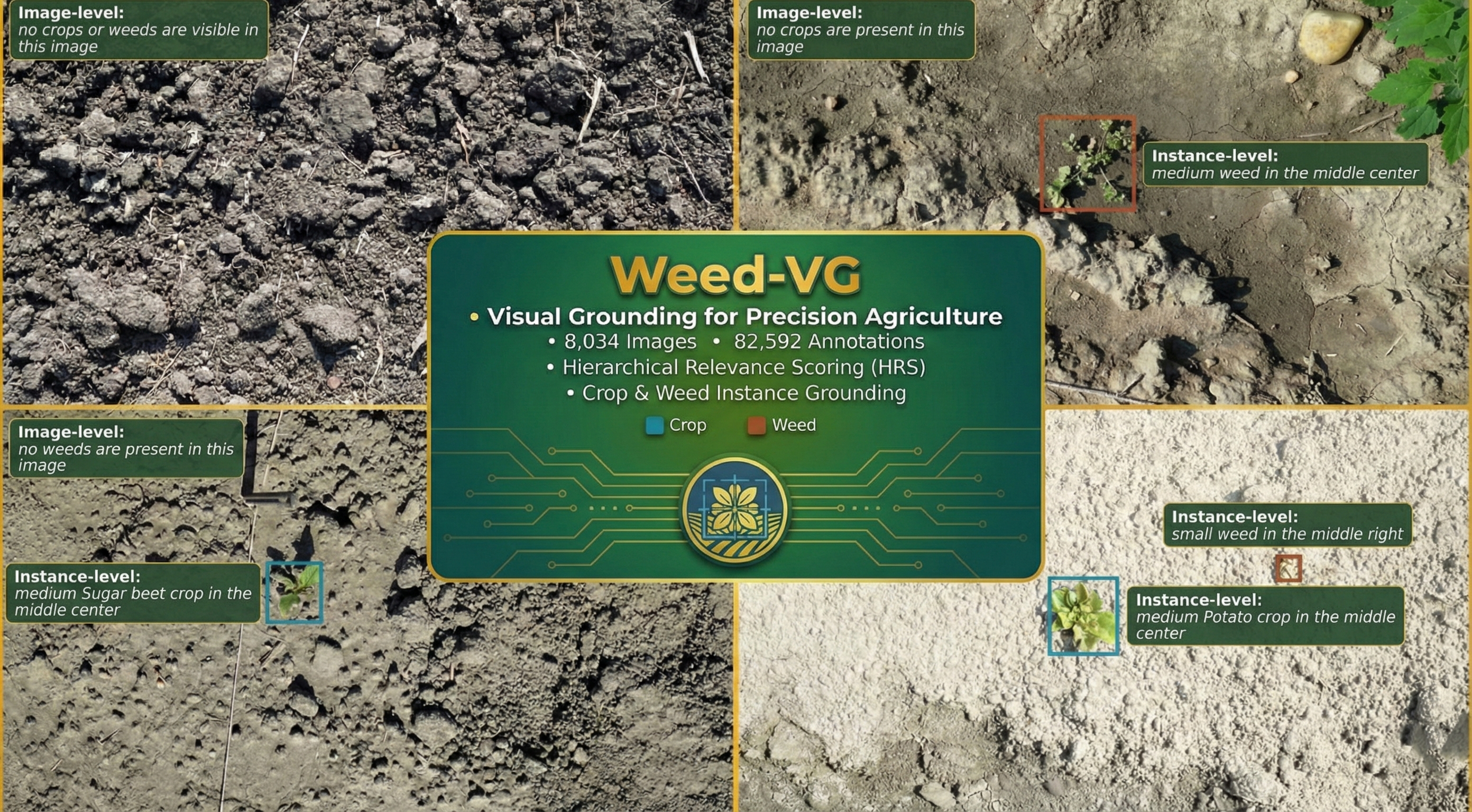}
    \caption{\textbf{Overview of the Weed-VG Framework for Precision Agriculture.}
    This figure summarises the paper’s main contributions for tackling the challenges of visual grounding (VG) in agricultural imagery. It presents gRef-CW, a generalised Referring-expression dataset for Crop and Weed visual grounding, comprising 8,000+ high-resolution field images and 82,000 annotations at both the image level and instance level. The figure also highlights Hierarchical Relevance Scoring (HRS), a modular method that enables existence-aware instance grounding, first determining whether the referred instance is present and then localising it.
    }
    \label{figure:ov}
\end{figure*}
\vspace{-7mm}
\begin{abstract}
Understanding field imagery such as detecting plants and distinguishing individual crop and weed instances, is a central challenge in precision agriculture. Despite progress in vision–language tasks like captioning and visual question answering, Visual Grounding (VG), localising language-referred objects, remains unexplored in agriculture. A key reason is the lack of suitable benchmark datasets for evaluating grounding models in field conditions, where many plants look highly similar, appear at multiple scales, and the referred target may be absent from the image.
To address these limitations, we introduce gRef-CW, the first dataset designed for generalised visual grounding in agriculture, including negative expressions. Benchmarking current state-of-the-art grounding models on gRef-CW reveals a substantial domain gap, highlighting their inability to ground instances of crops and weeds. Motivated by these findings, we introduce Weed-VG, a modular framework that incorporates multi-label hierarchical relevance scoring and interpolation driven-regression. 
Weed-VG advances instance-level visual grounding and provides a clear baseline for developing VG methods in precision agriculture. Code will be released upon acceptance.
\keywords{Precision Agriculture \and Generalised Referring Expression Comprehension \and Multi-modal Dataset \and Hierarchical Multi-label Learning} \vspace{-5mm}
\end{abstract}

\section{Introduction}
\label{sec:intro}
\vspace{-2mm}
Agriculture is essential for sustaining human life, and effective management of crops, weeds, diseases, and pests is critical for stable production \cite{azghadi2025precision,shinoda2025agrobench, zhou2024agribench, liu2024multimodal, arshad2025leveraging, quoc2025vision}. While VLMs have advanced tasks such as image captioning and Visual Question Answering (VQA) \cite{tumu2025exploring, li2024vision, dai2025improving}, VG—localising objects in images based on language queries—remains unexplored in agriculture \cite{xiao2024towards, pantazopoulos2025towards}. Unlike traditional object detectors, which are limited to predefined categories, the generalised form of VG (gVG) can handle multiple or absent targets and offer instance-aware localisation in complex field environments \cite{plummer2015flickr30k, krishna2017visual, guo2025visual, kuckreja2024geochat, schulter2023omnilabel, xie2023described}.

Crowded agricultural fields, with crops and weeds of varying sizes and absent targets, expose the limitations of standard VG models. Language-based queries offer flexible, context-aware specification, enabling monitoring of crop and weed presence, estimating quantities, and supporting selective weeding, fertilisation, irrigation, and harvesting. However, existing approaches \cite{dai2025improving, wang2025hierarchical, dai2025propvg} are still insufficient for real-world scenarios due to the domain shift. For instance, the well-known VG model GroundingDINO reports an mAP of 33.9 on agricultural images containing weeds \cite{robicheaux2025roboflow100}. Similarly, on the AgroBench dataset \cite{shinoda2025agrobench}, open-source VLMs perform close to random for weed identification, with the best models achieving only 55.17\% accuracy \cite{shinoda2025agrobench}. In addition, state-of-the-art grounding approaches such as InstanceVG \cite{dai2025improving} often fail to localise small objects. To systematically study these challenges, we introduce gRef-CW, the first agricultural VG dataset, designed with greater complexity and real-world variability compared to traditional VG benchmarks due to the following characteristics: \newline
\noindent\textbf{Novel and visually similar objects.}
In agricultural field images, VG is often challenged by objects that either fall outside the training distribution or closely resemble other instances \cite{madan2024revisiting}. Novel objects such as crops and weeds are unseen during training, and share overlapping features such as colour, shape, or texture \cite{robicheaux2025roboflow100}. Accurately distinguishing them is difficult when relying only on object-level features \cite{hao2025referring, guo2025beyond, yuan2021instancerefer, liu2025aerialvg, you2023ferret}. Therefore, effective detection must incorporate instance-level information, such as position and size, that could be encoded in language descriptions. \newline
\textbf{Instances of varying sizes.}
Agricultural images contain a range of instances, varying in size from small seedlings to larger crops or weeds. Smaller objects occupy fewer pixels, making detection and accurate prediction of the bounding box more challenging \cite{hemanthage2024recantformer, hu2023beyond, ma2024visual}. Instance-level detection should be performed for all objects that are distinguishable by human experts.

To address the above challenges, gRef-CW is designed to support instance-level VG in agricultural field imagery by providing multi-level textual annotations, bounding boxes, and segmentation masks. The dataset addresses the challenging task of distinguishing individual crops from weeds and determining their presence, considering significant intra-class variation. It contains 8,034 high-resolution images from CropAndWeed \cite{steininger2023cropandweed} with 78k crop and weed instances and 82k annotations, including 78,288 instance-level expressions and 4304 image-level expressions (see Fig. \ref{figure:ov}). This comprehensive annotation enables the development and evaluation of VLMs in field images, making it suitable for gVG tasks. Aiming to improve gVG models' performance on gRef-CW, we also design an innovative framework to leverage a text-conditioned proposal generator and decompose gVG into hierarchical global existence detection and instance relevance through a multi-label constraint-enforcing approach. We conduct extensive experiments to thoroughly validate the effectiveness of our proposed model, demonstrating its superior performance.
\vspace{-4mm}

\section{Related Work} \vspace{-1.5mm}
\label{sec:related_work}
\textbf{Vision-Language Models in Agriculture.}
In agriculture, VLMs have been applied to image captioning and VQA for applications including crop and plant health management as well as pest and insect detection \cite{haghighat2025multimodal,quoc2025vision, arshad2025leveraging, liu2024multimodal}, 
However, their grounding capability remains unexplored in this domain, despite grounding being a core requirement for fine grained vision language understanding. Strengthening grounding can also improve downstream performance, including VQA, by enabling precise referring and localisation \cite{reich2024uncovering}.
Existing agricultural multimodal benchmarks, including AgriBench \cite{zhou2024agribench}, AgroBench \cite{shinoda2025agrobench}, AgEval \cite{arshad2025leveraging}, CDDM \cite{liu2024multimodal}, AgMMU \cite{gauba2025agmmu}, and MIRAGE \cite{dongre2025mirage}, evaluate recognition and reasoning, but they are not designed for VG tasks. We address this gap by introducing a VG-specific agricultural dataset to leverage the power of VLMs in multimodal understanding, for instance-level crop and weed grounding.
\noindent
\\
\textbf{Generalised Visual Grounding.} 
gVG extends the classic VG task by addressing scenarios that involve zero or multiple targets. It can be further divided into two sub-tasks: generalised Referring Expression Comprehension (gREC)/Segmentation (gRES) \cite{liu2023gres, he2023grec}. Models have been developed to improve the results mainly in scenarios where object classes are known but are distinguished using the referring expressions \cite{ding2025multimodal, bai2025univg, cai2025naver, zeng2024investigating, chen2020cops}. 
These methods leverage Transformer-based architectures to directly predict referred targets. Domain-specific VG faces challenges such as tiny entities and object similarity in aerial and remote sensing imagery \cite{yuan2023rrsis, li2024language, lan2024language, liu2024rotated, wang2024multistage, hang2024regionally, ding2024visual, choudhury2025improving, ou2025geopix, zhang2025language, zhan2023rsvg} mainly due to the top-down perspective. Similarly, in medical imaging, grounding is difficult due to subtle boundaries and reliance on expert semantic cues rather than appearance alone \cite{huy2025seeing,he2025parameter,zou2024medrg,chen2023medical,luo2024vividmed}, making them distinct from natural scenes. Agricultural imagery presents analogous yet distinct challenges critical for gVG, where entity identification, such as detecting weeds and their growth stage, is complicated by high visual similarity to crops, specifically in crowded scenes. We leverage the gVG framework to address these fine-grained recognition challenges guided by language expressions, which remain unexplored in this domain.
\vspace{-4mm}

\section{Dataset} \vspace{-.5mm}
\label{sec:dataset}
The gRef-CW dataset comprises field images representing diverse field scenes with varying crop and weed densities. It is derived from the CropOrWeed9 subset of the CropAndWeed dataset \cite{steininger2023cropandweed}, with all labels mapped to eight crop types (Maize, Sugar beet, Bean, Pea, Sunflower, Soy, Potato, and Pumpkin) and one weed class. This subset was chosen because joint training across all crops with a unified weed category improves crop identification compared to single-crop models, enabling fine-grained crop–weed differentiation \cite{steininger2023cropandweed}. Each image includes manually refined bounding boxes and segmentation masks. To enhance annotation efficiency, images were first pre-segmented into soil and vegetation via colour-based thresholding, later replaced by a CNN-based segmentation model. Bounding boxes were generated from these masks and manually refined, with ambiguous or densely populated images validated through multi-annotator voting to ensure consistency and quality. The dataset captures real field variability, where crops and weeds look alike, especially at early stages, and correctly determining weed presence or absence is crucial for effective field management.
\begin{figure*}[!ht]
    \centering
    \includegraphics[width=0.74\linewidth]{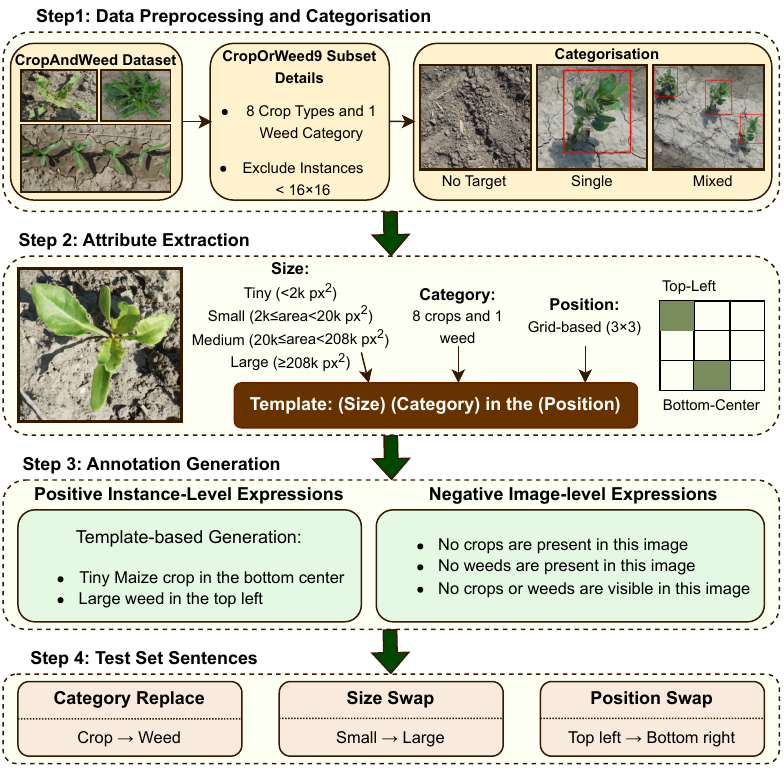}
    \caption{\textbf{gRef-CW Data Collection and Annotation Pipeline.} The dataset is constructed through a four-stage process: (1) Filtering out instances unrecognisable by humans and categorising images as single, mixed, or no-target from the selected subset; (2) Extracting instance attributes (e.g. size, category, and position) (3) Composing attributes into natural language templates to generate positive instance-level referring expressions; negative image-level sentences include the non-existence and (4) Replacing categories or swapping attributes to create test sentences.}
    \label{figure:co}
\end{figure*} \\
\textbf{Annotation pipeline.}
As shown in Fig.~\ref{figure:co}, we exclude instances smaller than 16×16 pixels, which are too small to be identified by humans. For each instance, we provide a referring expression that includes category, position, and size, along with a bounding box. Positions are extracted programmatically using a 3×3 grid based on bounding-box centre for each instance. Sizes are categorised into four levels—tiny, small, medium, and large—based on bounding-box area (Fig.~\ref{figure:co}).\\
\textbf{Dataset statistics.}
To improve robustness in category-absent scenarios, we generate negative referring expressions at both the image level (see the right box at step 3 in Fig.~\ref{figure:co}) and instance level expressions in the test set (Step 4) for the gRef-CW. This follows two strategies, Replace and Swap \cite{yang2025new, liu2024finecops}. Replace substitutes category of an instance referring expression, whereas Swap interchanges two attributes within the same category (see step 4 in Fig.~\ref{figure:co}). 
\begin{table*}[ht!]
\Huge
    \centering
    \setlength{\tabcolsep}{4pt} 
    \renewcommand{\arraystretch}{1.2} 
    \caption{Statistical comparison of classical referring expression datasets with gRef-CW. Instance and image dimensions are shown as square-root ranges. (Avg. Len. = average annotation length in words; Neg. Expr. = includes negative expressions.) This table contextualises the scale and complexity of gRef-CW rather than direct comparison of the annotation generation process, since our annotations are template-based, unlike the listed datasets.}
    \label{tab:rec-comparison}
    \resizebox{0.8\textwidth}{!}{%
    \begin{tabular}{l | rccr | cc | ccr}
    \specialrule{.1em}{.05em}{.05em}
    \multirow{3}{*}{\textbf{Dataset}} & \multicolumn{4}{c|}{\textbf{General Statistics}} & \multicolumn{2}{c|}{\textbf{Linguistics}} & \multicolumn{3}{c}{\textbf{Visual Properties}} \\
    \cline{2-10}
     & \multirow{2}{*}{Images} & \multirow{2}{*}{Instances} & \multirow{2}{*}{Anns} & \multirow{2}{*}{Cats} & Avg. & Neg. & Inst. & Norm. & Img. \\
     & & & & & Len. & Expr. & Size & Size & Size \\
    \hline
    RefCOCO \cite{yu2016modeling} & 19,994 & 50,000  & 142,210 & 80  & 3.49 & \xmark & 230--640  & 0.17--1.0 & 640 \\
    RefCOCO$+$ \cite{yu2016modeling} & 19,992 & 49,856  & 141,564 & 80  & 3.58 & \xmark & 230--640  & 0.17--1.0 & 640 \\
    RefCOCOg  \cite{nagaraja2016modeling}   & 26,711 & 54,822  & 104,560 & 80  & 8.46 & \xmark & 277--640  & 0.22--1.0 & 640 \\
    ReferItGame \cite{kazemzadeh2014referitgame} & 19,894 & 96,654  & 130,525 & 238 & 3.45 & \xmark & 360--480  & < 0.3$^{\dagger}$          & 480 \\
    \hline
    \rowcolor{gray!10} 
    \textbf{gRef-CW (Ours)} 
    & \textbf{8,034}  & \textbf{78,288} & \textbf{82,592} & \textbf{9} 
    & \textbf{6.34}   & \textbf{\cmark} 
    & \textbf{16--1,402} & \textbf{0.01--0.97} & \textbf{1,445} \\
    \specialrule{.1em}{.05em}{.05em}
    \end{tabular}
    }
    \vspace{-3mm}
    \begin{flushleft}
    \scriptsize{$^{\dagger}$ Minimum object size is not explicitly reported.}
    \end{flushleft}
    \vspace{-3mm}
\end{table*}

\begin{figure*}[!ht]
    \centering
    \includegraphics[width=1\linewidth]{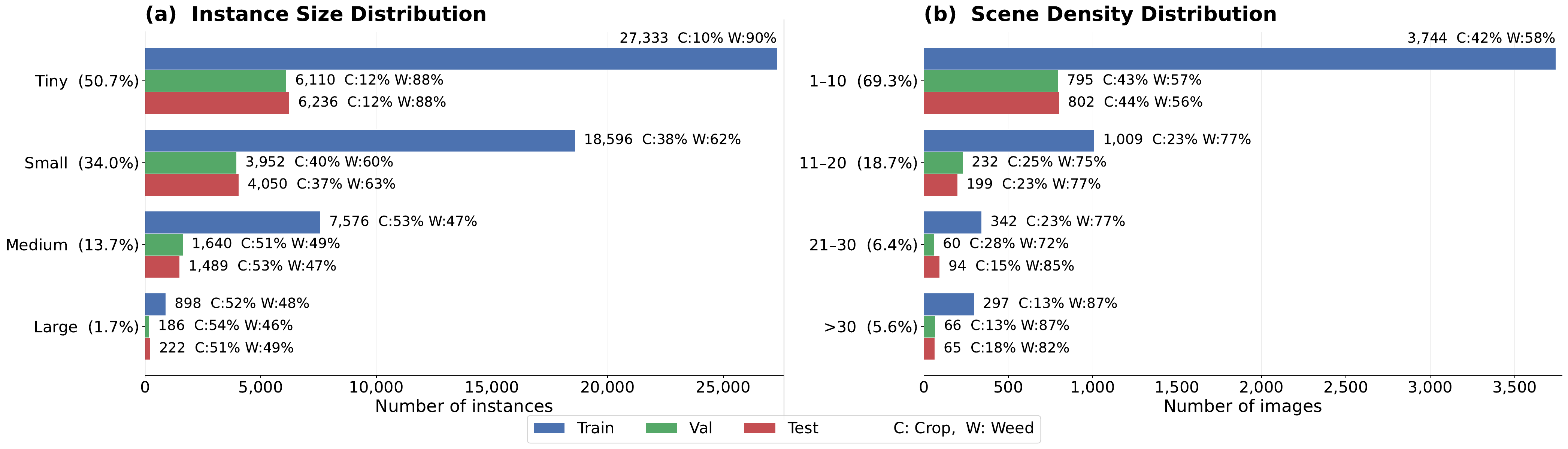}
    \caption{Detailed instance distribution of the gRef-CW across splits. The figure displays (a) the count of instances stratified by scale (from Tiny to Large) and (b) the number of images categorised by scene density. Annotations within the bars indicate the percentage of Crop (C) and Weed (W) for each subset.}
    \label{figure:stats}
\end{figure*}

To collect negative instance-level expressions in the test set, we randomly select one-third of the total 11,997 candidates for each change: category replacement, size swapping, and position swapping. This yields a total of 9,186 negative expressions (3,706 negative category, 3294 negative size, 2186 negative position), and the remaining are positive. All negatives are stored alongside positive expressions in a consistent annotation format, enabling evaluation under both object presence and absence conditions. The dataset is split into train, validation, and test sets with a ratio of 70:15:15, ensuring balanced representation of images containing only crops, only weeds, both, or neither.
An overview of gRef-CW statistics is provided in Table~\ref{tab:rec-comparison} compared to classical referring expression datasets. The Table shows gRef-CW's higher instance density, broader scale variation, instance-oriented expressions, and uniquely, the inclusion of negative expressions at both image and instance levels. As illustrated in Fig.~\ref{figure:stats}, gRef-CW presents a challenging long-tailed distribution where 84.7\% of instances are tiny or small, and scene complexity varies significantly from sparse to extremely dense clusters, where 30.7\% of images contain more than 10 instances.
\vspace{-8mm}

\section{Method} 
\begin{figure*}[!ht]
    \centering
    \includegraphics[width=0.71\linewidth]{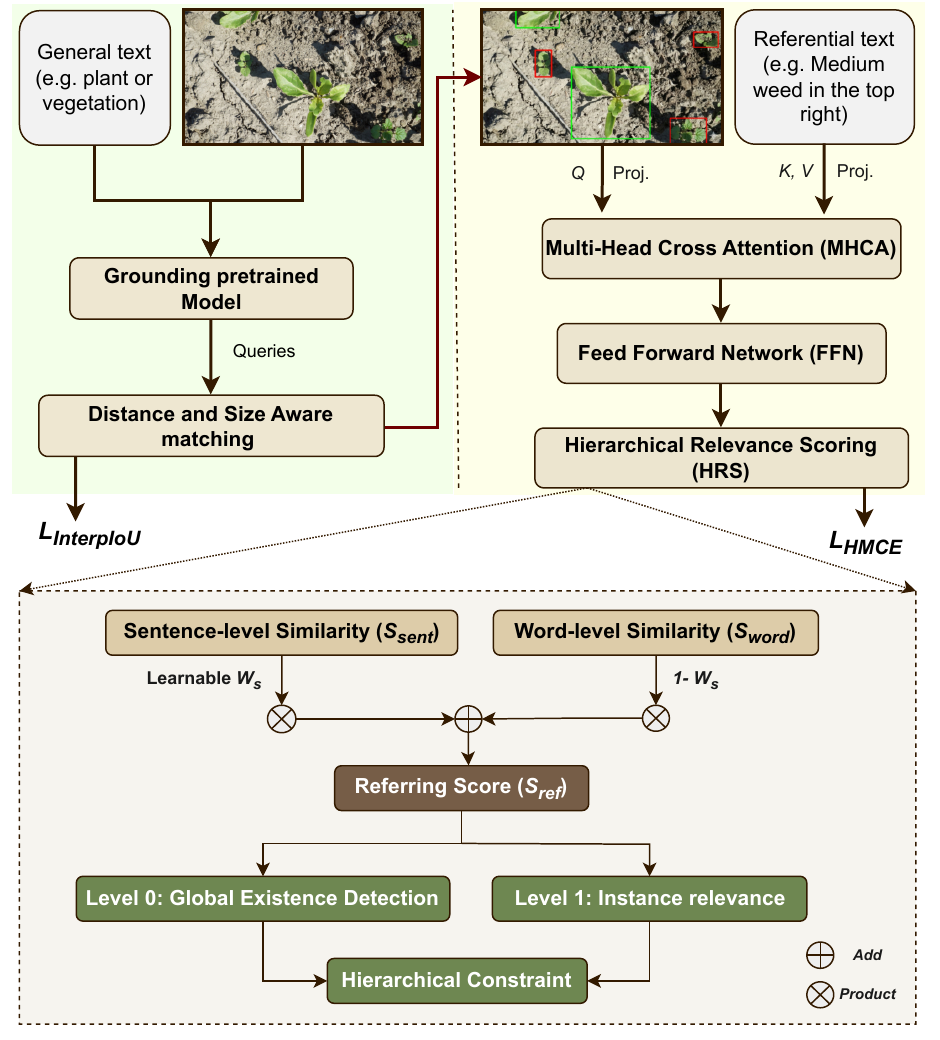}
    \caption{\textbf{Architecture of the Weed-VG framework.} It extends a grounding model with an HRS module that fuses visual and textual features via Multi-Head Cross-Attention and an FFN. HRS decomposes relevance into two levels: (Level 0) global existence detection, which predicts whether the referred object appears in the image, and (Level 1) instance relevance, which ranks region proposals by integrating sentence-level and word-level similarities. A constraint enforces logical consistency by conditioning instance localisation on global existence.}
    \label{figure:ar}
\end{figure*}

\vspace{-1.5mm}
\label{sec:method}
The proposed Weed-VG architecture follows a modular design, where a grounding model generates proposals, which are then refined by the HRS module that captures both image- and instance-level information across two label levels.
As shown in Fig.~\ref{figure:ar}, given an image and a general text input to the grounding model, we first perform distance- and size-aware matching over the proposals for regression. Instance-level expressions are then aligned with their corresponding proposals (queries) and, together with the text representations, projected into a shared embedding space. The projected features are then processed by Multi-Head Cross-Attention (MHCA) followed by a Feed-Forward Network (FFN) to produce aligned multimodal relevance logits, through which instance relevance is learned contrastively in HRS during training. HRS decomposes relevance into two levels: Level-0 global existence detection, which predicts whether the text refers to any object in the image, and Level-1 instance relevance, which scores each candidate region. Word-level and sentence-level similarities are computed and combined via a learnable weight to form a unified referring score. A hierarchical constraint, enforced through a hierarchical multi-level loss, is applied to instance-level predictions by the global existence score, ensuring logical consistency between levels. This modular design allows the model to jointly predict whether a referent exists and which specific instances correspond to the query, while remaining compatible with standard grounding inference pipelines.
\vspace{-5mm}
\subsection{Hierarchical Relevance Scoring with Constraint Enforcement}
\label{sec:hrs} 
\vspace{-2mm}
Given a textual query $T$ and a set of $N$ candidate regions $\mathcal{V}=\{v_j\}_{j=1}^{N}$ produced by the grounding model, HRS decomposes gVG into two levels. At Level~0 (\textit{Existence Detection}), the model predicts whether the query refers to any instance in the image. At Level~1 (\textit{Instance Discrimination}), it ranks candidate regions according to their relevance to the query. The final proposal score $s(v_j,T)\in\mathbb{R}$ is obtained by enforcing a hierarchical constraint that aligns instance-level relevance with the global existence prediction.

\textbf{Hierarchical Relevance Scoring.}
Our supervision strategy is structured across two hierarchical levels to ensure robust grounding.\\
\textit{Level~0: Existence Formulation.}
At the coarsest granularity, the model determines the global semantic state of the image. This is formulated as a multiclass classification task over a fixed vocabulary of image-level sentences. We define the pooled relevance logit $k$ by selecting the maximum proposal score for a given sentence, and compute the resulting image-level class probability as
\begin{equation}
    s_{\text{pool}}(k) = \max_{j} s(v_j, t_k), \quad P(k \mid I) = \text{Softmax}\big(s_{\text{pool}}(k)\big).
\end{equation}
where $I$ denotes the current image. Optimisation at this stage employs a cross-entropy loss $L_{\text{lvl0}}$, ensuring the model verifies global existence before performing instance-level scoring.
\textit{Level~1: Instance Relevance Formulation.}
Conditioned on the global existence prediction, the model proceeds to rank valid object instances. We optimise this ranking using a Binary Cross-Entropy (BCE) loss, \(L_{\text{lvl1}}\), which accommodates both single-positive and multi-positive scenarios. The instance-level relevance logits are derived via a contrastive formulation that synergises sentence-level and word-level query-to-text similarities, fused by $k$ to produce the $S_{\text{ref}}$. 

Sentence-level textual features \(f_s\) are extracted by applying valid-mask max pooling over word-level embeddings \(f_t\); this operation preserves the maximum response of each token across channels by selecting the highest value along the token dimension. Proposal-to-sentence similarity is then computed using a learnable temperature-scaled cosine similarity initialised to 0.07. Concurrently, the word-level branch computes a similarity matrix \(S_{\text{word}} \in \mathbb{R}^{N \times N_t}\) between proposal features and individual word tokens. 

To dynamically balance these contributions, a learnable weight \(w_s\) is generated by applying a multi-layer perceptron followed by a sigmoid activation to the global textual feature \(f_s\). The referring score for each proposal is computed as:
\begin{equation}
S_{\text{ref}} = w_s \cdot S_{\text{sent}} + (1 - w_s) \cdot \text{MaxPool}(S_{\text{word}}).
\end{equation}
This formulation adaptively integrates sentence-level semantics with fine-grained word-level cues, enabling robust instance discrimination.

\textbf{Hierarchical Constraint Enforcement.} A key logical dependency is that an instance cannot be localised if it does not exist in the image. We enforce this by applying a constraint on the instance-level loss using the image-level loss as a lower bound in the hierarchy: $L_{\text{lvl1}}^{\text{constrained}} = \max\big(L_{\text{lvl1}}, L_{\text{lvl0}}\big)$.
This max operator guarantees that when global existence recognition is poor, instance discrimination cannot be performed. This effectively postpones $L_{\text{lvl1}}$ until the existence prediction has stabilised, and leads to training dynamics that translate into a proposal-ranking behaviour at inference time.
\vspace{-4mm}
\subsection{IoU-Driven Interpolation for Bounding Box Regression}
\vspace{-1mm}
Crop and weed visual grounding poses challenges due to extreme scale variation depending on the growth stage, with object areas ranging from as little as \(0.01\%\) to \(0.97\%\) of the image. Direct application of standard IoU-based regression losses under such conditions often leads to unstable gradients and slow convergence \cite{liu2025interpiou}. To mitigate this issue, we adopt an IoU-driven interpolation strategy, termed InterpIoU, which stabilises bounding box optimisation without introducing handcrafted geometric penalties. Given a predicted bounding box \( B_{\text{pred}} \) and its corresponding ground truth \( B_{\text{gt}} \), an intermediate box is constructed by linear interpolation,
\begin{equation}
B_{\text{int}} = (1 - \alpha) B_{\text{pred}} + \alpha B_{\text{gt}}, \qquad 0 < \alpha < 1.
\label{eq:interp_box}
\end{equation}
The regression objective then combines the standard IoU loss with an auxiliary IoU term computed using the interpolated with $\alpha = 0.99$ to address extreme scale variation following the original implementation \cite{liu2025interpiou},
\begin{equation}
L_{\text{InterpIoU}}(B_{\text{pred}}, B_{\text{gt}}) =
L_{\text{IoU}}(B_{\text{pred}}, B_{\text{gt}})
+
\, L_{\text{IoU}}(B_{\text{int}}, B_{\text{gt}}),
\end{equation}
where \( L_{\text{IoU}} = 1 - \text{IoU} \). The auxiliary term provides smooth, non-zero gradients even when the predicted box is poorly aligned with the ground truth, effectively guiding predictions toward the target while preserving geometric integrity. This formulation is more robust to variations in plant instance scale and density across images.

\textbf{Distance and Size Aware Matching.}
In the specific implementation, the matching cost is formulated to jointly account for overlap, centre distance, and relative scale differences between proposals and ground-truth boxes. This design encourages assignments that are consistent in position, size, and spatial alignment. The matching cost $C_{ij}$ between the $i$-th proposal $P_i$ and the $j$-th ground-truth instance $G_j$ is defined as:
\begin{equation}
\resizebox{0.9\textwidth}{!}{$
C_{ij} = (1 - \text{IoU}(P_i,G_j)) + \lambda_{\text{centre}} \|\mathbf{c}(P_i) - \mathbf{c}(G_j)\|^2 + \lambda_{\text{size}} \left( \frac{|w_{P_i} - w_{G_j}|}{w_{G_j}} + \frac{|h_{P_i} - h_{G_j}|}{h_{G_j}} \right)
$}
\end{equation}
\noindent where $\mathbf{c}(P_i)$ and $\mathbf{c}(G_j)$ denote the centres of the predicted and ground-truth boxes, respectively. In this equation, the first term measures the standard IoU overlap, the second term penalises the squared $L_2$ distance between box centres, and the third term captures the relative discrepancy in width and height. The weighting parameters are empirically chosen to prioritise box-centre alignment ($\lambda_{\text{centre}} = 2.0$) while remaining size-aware ($\lambda_{\text{size}} = 0.5$) due to extreme scale variation that causes the domination of centre error.
\vspace{-4mm}
\subsection{Loss Function}
\vspace{-2mm}
The final training objective integrates hierarchical semantic alignment and robust spatial localisation through a unified loss formulation:
\begin{equation}
L_{\text{total}} = L_{\text{HMCE}} + L_{\text{InterpIoU}},
\end{equation}
where $L_{\text{HMCE}}$ is the Hierarchical Multi-label Constraint Enforcement loss, and $L_{\text{InterpIoU}}$ is the IoU-driven interpolation loss for bounding box regression.

Specifically, the hierarchical loss is defined as
\begin{equation}
L_{\text{HMCE}} = 
\lambda_0 \, L_{\text{lvl0}}
+ \lambda_1 \, L_{\text{lvl1}}^{\text{constrained}}.
\end{equation}
The adaptive weights $\lambda_0$ and $\lambda_1$ are chosen based on image type: for multi-referent (mixed) images, $(1.0, 2.5)$ to prioritise instance discrimination; for single-category images, $(1.0, 2.0)$; and for empty images, $(1.0, 0)$ to focus solely on existence detection.
\vspace{-4mm}

\section{Experiments} \vspace{-2mm}
\label{sec:exp}
\subsection{Implementation Details} \vspace{-1mm}
The proposed framework modularises grounding frameworks. Specifically, we employ GroundingDINO \cite{liu2024grounding} with a Swin Transformer visual backbone and a BERT-base text encoder in our experiments for proposal generation using the general text mentioned in Fig.~\ref{figure:ar}.
We then perform training in two stages to avoid instability observed in end-to-end joint training, which caused a reduction in regression performance. In the first stage, the model is initialised from the original checkpoint, and only the last decoder layer and box head are fine-tuned using the InterpIoU loss for 100 epochs. In the second stage, training is restricted to the projection/attention layers and the HRS module for 60 epochs using two-level labels, focusing on learning hierarchical semantics and instance-level distinctions. The epoch schedule is empirically selected to balance convergence and overfitting.
Training batches are carefully constructed to include both positive instance-level expressions and negative sentences, including no-target and absence queries. A batch size of 4 is used, accounting for the four image types. Since annotations do not include negative sentences for images containing both crops and weeds, and hierarchical labels cannot be empty, we introduce a special token \texttt{[EMPTY]} to handle these cases. The maximum encoder input length for testing is set to 64 tokens. Optimisation is performed using AdamW with a cosine annealing learning rate schedule initialised at $2\times10^{-4}$.
The data augmentation pipeline incorporates Copy–Paste augmentation to increase instance diversity, as well as random rotations and colour jittering to simulate variations in viewpoint, illumination, and plant appearance. All experiments were conducted using an NVIDIA A100 GPU on the Bunya HPC cluster and an NVIDIA RTX PRO 6000 Blackwell GPU.
\vspace{-5mm}
\subsection{Evaluation Metric} \vspace{-2mm} 
We evaluate four VG baselines on gRef-CW using Recall@0.5, Top-k Accuracy, and mIoU to assess retrieval, ranking, and localisation and to characterise dataset challenges. Since VLMs perform near random chance on agricultural tasks such as crop and weed detection \cite{shinoda2025agrobench}, and grounding models cannot be fine-tuned on multi-level labels without HRS, direct comparison of the methodological improvements introduced by HRS relative to other methods is avoided to ensure fairness. Instead, Weed-VG results are reported by modularising a GroundingDINO \cite{liu2024grounding} baseline within our framework (Table~\ref{tab:main_results_val_test}).

Due to the baselines' high instance miss on gRef-CW because of the sensitivity to threshold selection, we re-implement baselines to evaluate \textit{all} their proposals given an instance sentence, and introduce Negative Accuracy (Neg-Acc), a threshold-free GIoU-based \cite{rezatofighi2019generalized} metric, evaluated on sentences where no target exists. We report Neg-Acc alongside Recall to account for differences in proposal counts across baselines. A prediction is correct if its maximum GIoU with any ground truth is $\leq 0$. Intuitively, Neg-Acc measures the model’s ability to correctly refer to background. Evaluated together with Recall, it acts as a test of conditional understanding. Ideally, a model should both detect targets when present (High Recall) and abstain when absent (High Neg-Acc). Existing baselines fail this joint test: they achieve moderate detection only by flooding the scene with proposals, but their Neg-Acc collapse to ${\sim}3$--$7.5\%$ (SAM3 ${\sim}25\%$).

Top-1 measures the fraction of queries where the highest-ranked proposal reaches IoU $\geq0.5$ with the ground-truth. High Top-5 but low Top-1 indicates the model finds plausible candidates but fails to rank the target correctly. To analyse language understanding in detail, metrics are further stratified by instance category, attribute type, and scene density (Tables~\ref{tab:detailed_performance_tabu} and \ref{tab:complexity_tabu_style}).

\vspace{-4mm}
\subsection{Results} \vspace{-1mm}
\textbf{Main Benchmark Analysis.} Table~\ref{tab:main_results_val_test} illustrates baselines' dominant failure modes: weak instance ranking and localisation as well as insufficient recall. 
\begin{table}[ht!]
\Huge
    \centering
    \setlength{\tabcolsep}{4pt}
    \renewcommand{\arraystretch}{1.3}
    \caption{\textbf{Main results on the gRef-CW.} We report Top-1, Top-5, R@0.5, and mIoU on both the Validation and Test sets. Negative Accuracy (Neg-Acc) is evaluated on the instance-level negative sentences provided in the Test set.}
    \label{tab:main_results_val_test}
    \resizebox{0.75\linewidth}{!}{%
    \begin{tabular}{l|cccc|ccccc}
    \specialrule{.1em}{.05em}{.05em}
    \multirow{2}{*}{\textbf{Model}} & \multicolumn{4}{c|}{\textbf{Validation}} & \multicolumn{5}{c}{\textbf{Test}} \\
    \cline{2-10}
     & Top-1 & Top-5 & R@0.5 & mIoU & Top-1 & Top-5 & R@0.5 & mIoU & Neg-Acc \\
    \hline
    MDETR \cite{kamath2021mdetr}  & 10.20 & 12.81 & 7.82 & \underline{52.35} & 10.16 & 12.97 & 7.78 & \underline{54.19} & 3.32 \\
    GDINO-T \cite{liu2024grounding} & 12.89 & 33.04 & 21.50 & 18.93 & 11.92 & 31.99 & 20.34 & 17.44 & 2.88 \\
    GDINO-L \cite{liu2024grounding} & 19.74 & 42.58 & 28.16 & 23.08 & 20.38 & 43.49 & 28.73 & 23.68 & 7.52 \\
    SAM3 \cite{carion2025sam}  & \underline{33.85} & \underline{66.21} & \underline{46.87} & 31.98 & \underline{34.88} & \underline{66.80} & \underline{46.65} & 32.76 & \underline{25.53} \\
    \hline
    \rowcolor{gray!10} \textbf{Weed-VG (Ours)} & \textbf{63.01} & \textbf{81.64} & \textbf{55.71} & \textbf{58.12} & \textbf{62.42} & \textbf{82.45} & \textbf{55.44} & \textbf{57.25} & \textbf{78.35} \\
    \specialrule{.1em}{.05em}{.05em}
    \end{tabular}
    }
\end{table}
MDETR \cite{kamath2021mdetr} achieves relatively high localisation quality, yet its Top-1 accuracy remains near 10\% and R@0.5 below 8\%. The large gap between mIoU ($\approx 54$) and R@0.5 ($\approx 7.8$) indicates that when a correct instance is retrieved, the box is spatially accurate; however, the model rarely retrieves the correct instance. This suggests that agricultural referring is constrained by proposal recall under extreme small-object conditions.
In contrast, SAM3 \cite{carion2025sam} dramatically increases R@0.5 to 46.87\% (Val) and 46.65\% (Test), confirming higher coverage through dense proposal generation. However, its Top-1 accuracy remains limited to 33.85\% / 34.88\%. The 12-point gap between R@0.5 ($\approx 46.7$) and Top-1 ($\approx 34.9$) indicates that although the correct instance is often present among candidates, ranking fails in roughly one out of three cases. This behaviour is further reflected in mIoU (31.98 / 32.76), which remains substantially lower than MDETR \cite{kamath2021mdetr} despite far higher recall, showing that proposal abundance alone does not guarantee precise localisation.
Furthermore, increasing GroundingDINO backbone's capacity improves Test Top-1 accuracy, yet it still lags behind SAM3 \cite{carion2025sam}, indicating that scaling the detector alone cannot resolve instance ambiguity.
After integrating GroundingDINO \cite{liu2024grounding} in Weed-VG, Top-1 accuracy increases to 63.01\% (Val) and 62.42\% (Test). Importantly, this improvement does not sacrifice recall: R@0.5 increases further to 55.71\% / 55.44\%. Simultaneously, mIoU rises to 58.12\% / 57.25\%. The joint increase across Top-1, R@0.5, and mIoU indicates that Weed-VG resolves both retrieval and ranking failures rather than improving a single metric dimension.

\begin{figure*}[ht!]
    \centering
    \includegraphics[width=1\linewidth]{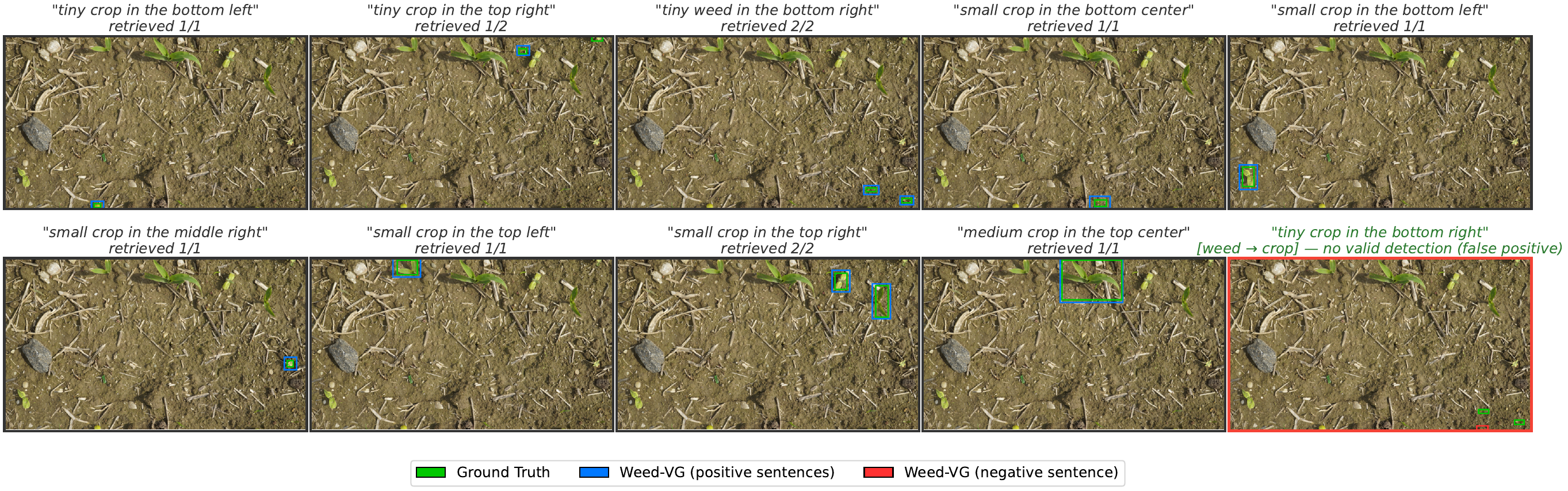}
    \caption{Qualitative results of Weed-VG on positive and negative referring expressions in gRef-CW across challenging scales (tiny–medium).}
    \label{figure:vis}
\end{figure*}

\textbf{Scale-Specific Behaviour.} Table~\ref{tab:detailed_performance_tabu} shows that the baselines exhibit severe collapse on tiny and small instances which are dominant in gRef-CW. On the Test set, both GDINO-T and GDINO-L \cite{liu2024grounding} perform poorly on tiny instances, with Top-1 accuracies below 2\%, highlighting failure under extreme scale reduction. SAM3 \cite{carion2025sam} improves tiny crop accuracy to 13.42\%, but this is still far below its 71.33\% performance on large crops, revealing a substantial 58-point gap due to scale sensitivity. Weed-VG reduces this disparity substantially with the integration of GDINO-L. The scale gap between tiny and large crops shrinks to only 17 points (54.66 to 71.90). This confirms that hierarchical scoring stabilises ranking under extreme scale compression. Notably, on large crops GDINO-L achieves 74.33\% Top-1, slightly higher than Weed-VG’s 71.90\%. However, the advantage of Weed-VG emerges in balancing the results across scales. This suggests that the contrastive relevance scoring, utilising both word- and sentence-level text processing while being existence-aware due to the hierarchy, highly benefits target referral.
\begin{table*}[ht!]
\huge
    \centering
    \setlength{\tabcolsep}{3pt} 
    \renewcommand{\arraystretch}{1.3} 
    \caption{Detailed performance breakdown by instance scale on the gRef-CW dataset. We report Top-1 accuracy and mIoU for crops and weeds separately on both Validation and Test sets. \textbf{Best results} are in bold, \underline{second-best} are underlined.}
    \label{tab:detailed_performance_tabu}
    \resizebox{0.96\textwidth}{!}{%
    \begin{tabular}{l|c|c|c|c|c||c|c|c|c}
    \specialrule{.1em}{.05em}{.05em}
    \multicolumn{1}{c|}{\multirow{2}{*}{\textbf{Method}}} & \multirow{2}{*}{\textbf{Set}} & 
    \multicolumn{4}{c||}{\textbf{Crop (Top-1 | mIoU)}} & 
    \multicolumn{4}{c}{\textbf{Weed (Top-1 | mIoU)}} \\
    \cline{3-10}
     & & Tiny & Small & Med & Large 
     & Tiny & Small & Med & Large \\
    \hline
    \multirow{2}{*}{MDETR \cite{kamath2021mdetr}}
     & \huge{Val}  & —$^{\dagger}$ & 4.17 | 22.84 & 10.28 | 52.31 & 11.29 | 60.29 & —$^{\dagger}$ & 2.08 | 19.27 & 10.12 | 50.80 & 12.91 | 69.14 \\
     & \huge{Test} & —$^{\dagger}$ & 3.31 | 21.18 & 10.34 | 53.56 & 11.71 | 63.40 & —$^{\dagger}$ & 2.79 | 17.61 & 9.68 | 52.55 & 13.09 | 71.48 \\
    \hline
    \multirow{2}{*}{GDINO-T \cite{liu2024grounding}}
     & \huge{Val}  & 0.68 | 0.74 & 11.83 | 14.87 & 23.36 | 35.38 & 35.98 | 60.04 & 0.00 | 0.26 & 6.50 | 9.03 & 18.29 | 28.72 & 43.18 | 73.82 \\
     & \huge{Test} & 0.00 | 0.05 & 7.56 | 9.82 & 23.75 | 34.86 & 31.64 | 53.53 & 0.31 | 0.72 & 6.45 | 8.75 & 20.79 | 31.48 & 43.32 | 68.73 \\
    \hline
    \multirow{2}{*}{GDINO-L \cite{liu2024grounding}}
     & \huge{Val}  & 0.70 | 2.21 & 22.48 | 23.24 & 41.52 | 50.46 & 55.99 | \textbf{70.63} & 0.75 | 1.29 & 12.44 | 13.85 & 37.15 | 45.26 & 63.53 | \textbf{79.55} \\
     & \Huge{Test} & 1.67 | 1.94 & 21.42 | 22.18 & 49.14 | 58.36 & 59.44 | \textbf{74.33} & 0.83 | 1.53 & 12.27 | 13.70 & 37.30 | 44.94 & 65.20 | \textbf{82.30} \\
    \hline
    \multirow{2}{*}{SAM3 \cite{carion2025sam}}
     & \huge{Val}  & \underline{9.64} | \underline{8.77} & \underline{44.28} | \underline{42.10} & \underline{57.16} | \underline{56.87} & \underline{64.90} | 65.07 & \underline{16.38} | \underline{14.09} & \underline{36.45} | \underline{33.86} & \underline{55.43} | \underline{53.95} & \textbf{76.85} | \underline{77.52} \\
     & \huge{Test} & \underline{13.42} | \underline{11.84} & \underline{48.63} | \underline{46.58} & \underline{61.20} | \underline{60.38} & \underline{71.33} | \underline{70.58} & \underline{16.17} | \underline{13.97} & \underline{35.57} | \underline{32.53} & \underline{55.12} | \underline{53.43} & \textbf{78.62} | \underline{78.30} \\
    \hline
    \rowcolor{gray!10} 
    \textbf{Weed-VG} & \huge{Val}  & \textbf{59.35} | \textbf{52.95} & \textbf{70.12} | \textbf{64.35} & \textbf{71.59} | \textbf{69.04} & \textbf{71.87} | \underline{70.43} & \textbf{57.25} | \textbf{50.44} & \textbf{59.18} | \textbf{55.28} & \textbf{67.93} | \textbf{65.70} & \underline{72.15} | 70.38 \\
    \rowcolor{gray!10} 
    \textbf{(Ours)} & \huge{Test} & \textbf{54.66} | \textbf{50.28} & \textbf{64.78} | \textbf{61.33} & \textbf{70.92} | \textbf{66.16} & \textbf{71.90} | 68.70 & \textbf{53.44} | \textbf{47.70} & \textbf{55.99} | \textbf{53.19} & \textbf{66.42} | \textbf{64.48} & \underline{76.79} | 70.69 \\
    \specialrule{.1em}{.05em}{.05em}
    \end{tabular}
    }
    \vspace{-2mm}
    \begin{flushleft}
    \scriptsize{$^{\dagger}$ MDETR \cite{kamath2021mdetr} produces no valid detections at this scale.}
    \end{flushleft}
    \vspace{-2mm}
\end{table*}
\\ \textbf{Scene Density Robustness.} Scene density further exposes ranking instability. As shown in Table~\ref{tab:complexity_tabu_style}, for crops in sparse scenes (1–10 instances), SAM3 \cite{carion2025sam} achieves 49.58 mIoU (Test). When density exceeds 30 instances, its mIoU drops to 25.81, a 23.77-point degradation. R@0.5 also decreases from 84.77 to 31.26, indicating severe interference among proposals. Robust performance in dense agricultural scenes, with more than 30 instances per image, is achieved by the proposed interpolation-based box regression enhanced with distance- and size-aware matching strategy in our framework, resulting in crop mIoU of 47.58 and weed mIoU remaining at 50.00.
\begin{table*}[ht!]
    \Huge
    \centering
    \setlength{\tabcolsep}{3pt} 
    \renewcommand{\arraystretch}{1.2} 
    \caption{Scalability with respect to scene density on the gRef-CW dataset. We report R@0.5 and mIoU for crops and weeds on both Validation and Test sets, grouped by number of instances in the scene. \textbf{Best results} are in bold, \underline{second-best} are underlined.}
    \label{tab:complexity_tabu_style}
    \resizebox{\textwidth}{!}{%
    \begin{tabular}{ll | c|c|c|c| >{\columncolor{gray!10}}c || c|c|c|c| >{\columncolor{gray!10}}c}
    \specialrule{.1em}{.05em}{.05em}
    \multicolumn{2}{l|}{\multirow{2}{*}{\textbf{Scene Density}}} & 
    \multicolumn{5}{c||}{\textbf{Crop (R@0.5 | mIoU)}} &
    \multicolumn{5}{c}{\textbf{Weed (R@0.5 | mIoU)}} \\
    \cline{3-12}
     & & MDETR \cite{kamath2021mdetr} & GDINO-T \cite{liu2024grounding} & GDINO-L \cite{liu2024grounding} & SAM3 \cite{carion2025sam} & \textbf{Weed-VG (Ours)}
     & MDETR \cite{kamath2021mdetr} & GDINO-T \cite{liu2024grounding} & GDINO-L \cite{liu2024grounding} & SAM3 \cite{carion2025sam} & \textbf{Weed-VG (Ours)} \\
    \hline
    \multicolumn{12}{l}{\Huge\textit{1--10 Instances}} \\
    \midrule
     & Val  & 22.03 | \underline{54.84} & 58.12 | 29.61 & 68.53 | 37.80 & \textbf{85.48} | 44.72 & \underline{84.26} | \textbf{66.80} & 10.14 | \underline{58.46} & 30.41 | 24.29 & 39.55 | 26.80 & \underline{69.35} | 37.67 & \textbf{69.73} | \textbf{58.75} \\
     & Test & 21.94 | \underline{57.76} & 51.09 | 24.37 & 67.93 | 38.67 & \textbf{84.77} | 49.58 & \underline{82.85} | \textbf{63.92} & 10.83 | \textbf{59.77} & 31.17 | 25.83 & 39.88 | 27.65 & \textbf{69.37} | 37.63 & \underline{69.02} | \underline{56.36} \\
    \hline
    \multicolumn{12}{l}{\Huge\textit{11--20 Instances}} \\
    \midrule
     & Val  & 14.85 | \underline{48.33} & 40.86 | 18.37 & 51.43 | 29.00 & \underline{67.70} | 42.07 & \textbf{73.63} | \textbf{62.56} & 3.75 | \underline{46.43} & 12.00 | 10.56 & 18.68 | 12.33 & \underline{41.23} | 23.85 & \textbf{53.77} | \textbf{53.36} \\
     & Test & 16.69 | \underline{43.90} & 39.88 | 15.77 & 58.21 | 30.52 & \underline{69.28} | 41.57 & \textbf{74.59} | \textbf{60.35} & 3.71 | \textbf{52.95} & 11.48 | 11.93 & 18.19 | 13.32 & \underline{39.84} | 22.37 & \textbf{54.64} | \underline{50.50} \\
    \hline
    \multicolumn{12}{l}{\Huge\textit{21--30 Instances}} \\
    \midrule
     & Val  & 8.59 | \underline{50.97} & 13.79 | 14.99 & 18.38 | 19.56 & \underline{30.55} | 32.66 & \textbf{44.15} | \textbf{58.66} & 3.23 | \underline{44.21} & 9.49 | 7.52 & 14.61 | 12.45 & \underline{32.64} | 21.45 & \textbf{47.91} | \textbf{54.58} \\
     & Test & 12.50 | \underline{54.92} & 28.89 | 13.54 & 38.69 | 29.54 & \underline{52.08} | 41.51 & \textbf{60.42} | \textbf{57.87} & 3.52 | \underline{49.55} & 9.07 | 8.64 & 15.23 | 11.15 & \underline{31.13} | 18.92 & \textbf{46.82} | \textbf{51.19} \\
    \hline
    \multicolumn{12}{l}{\Huge\textit{$>$30 Instances}} \\
    \midrule
     & Val  & 4.26 | \underline{42.02} & 17.71 | 12.22 & 23.99 | 18.13 & \underline{36.10} | 21.33 & \textbf{50.67} | \textbf{53.26} & 2.20 | \underline{47.09} & 4.33 | 6.36 & 6.88 | 9.57 & \underline{18.08} | 16.65 & \textbf{32.10} | \textbf{50.11} \\
     & Test & 4.75 | \underline{40.33} & 12.80 | 9.09 & 20.11 | 16.40 & \underline{31.26} | 25.81 & \textbf{44.42} | \textbf{47.58} & 0.94 | \underline{41.58} & 4.08 | 5.15 & 7.53 | 8.60 & \underline{18.93} | 19.57 & \textbf{31.71} | \textbf{50.00} \\
    \specialrule{.1em}{.05em}{.05em}
    \end{tabular}
    }
\end{table*} \\
\textbf{Negative Expression Handling.} As shown in Table~\ref{tab:nacc_structured}, standard grounding pipelines tend to propose regions that have overlap with other ground-truth boxes as quantified by Neg-Acc in Table~\ref{tab:nacc_structured}. SAM3 \cite{carion2025sam} improves Neg-Acc to 25.53\%, largely due to its proposal diversity, yet still fails in nearly three out of four negative cases. Dominant negative accuracy, with a consistent 78.35\% average Neg-Acc across manipulation types, is achieved by the proposed Level-0 existence detection enforced by hierarchical constraints in our framework, resulting in the explicit determination of referent existence and the generation of valid background proposals instead of referring to other ground-truth instances as shown in Fig.~\ref{figure:vis} (red box).
\FloatBarrier
\begin{wraptable}{r}{0.41\columnwidth} 
    \Huge
    \centering
    \setlength{\tabcolsep}{3pt} 
    \renewcommand{\arraystretch}{1.2} 
    \caption{Negative Accuracy on the gRef-CW Test set. We report Neg-Acc for Replace and Swap changes, together with the weighted average across categories. \textbf{Best results} are in bold, \underline{second-best} are underlined.}
    \label{tab:nacc_structured}
    \resizebox{\linewidth}{!}{
    \begin{tabular}{l | c | cc | c}
    \specialrule{.1em}{.05em}{.05em}
    \multirow{3}{*}{\textbf{Method}} & \multicolumn{4}{c}{\textbf{Neg-Acc}} \\
    \cline{2-5}
     & \textbf{Replace} & \multicolumn{2}{c|}{\textbf{Swap}} & \multirow{2}{*}{\textbf{Avg.}} \\
    \cline{2-2} \cline{3-4}
     & Category & Size & Position & \\
    \hline
    MDETR \cite{kamath2021mdetr}           &  3.26 &  4.91 & 54.55 & 3.32 \\
    GDINO-T \cite{liu2024grounding}         & 14.62 & 11.32 & 16.82 & 2.88 \\
    GDINO-L \cite{liu2024grounding}         & 12.84 & 13.36 & 42.42 & 7.52 \\
    SAM3 \cite{carion2025sam}            & \underline{41.02} & \underline{34.03} & \underline{60.75} & \underline{25.53} \\
    \hline
    \rowcolor{gray!10}
    \textbf{Weed-VG (Ours)} & \textbf{79.15} & \textbf{77.77} & \textbf{78.01} & \textbf{78.35} \\
    \specialrule{.1em}{.05em}{.05em}
    \end{tabular}
    }
\end{wraptable}
\FloatBarrier
\textbf{Qualitative Results.}
As shown in Fig.~\ref{figure:vis}, the framework demonstrates proper instance recall and localisation. It also effectively rejects absent crop instances of the same size and spatial location by referring to background regions. Notably, when the category changes but the size and location remain similar, Weed-VG avoids making false positives by referring to the same instances (see bottom right, first image). However, limitations remain in highly ambiguous scenarios. The model occasionally suffers from partial recall as shown in the second image, top right. Furthermore, extreme visual similarity between early-stage "tiny" plants can cause category confusion, overriding both instance referral and the rejection mechanism.
\vspace{-1mm}
\vspace{-7mm}
\subsection{Ablations}
\label{sec:ablations}
\vspace{-1mm}
To analyse the contribution of the components in the proposed framework, we conduct an ablation study as shown in Table~\ref{tab:ablation_hrs_clean}. Using only sentence-level similarity drops Top-1 by 13.13 and mIoU by 7.9, while word-only cues degrade further to 36.19\% Top-1, showing that grounding requires joint sentence semantics and token alignment. Removing Query Projection causes the largest collapse for Top-1 and mIoU, indicating that effective instance grounding highly depends on projecting visual and textual features into a shared learnable space that jointly captures the context of both image-level and instance-level expressions hierarchically.
Removing the Hierarchical Constraint sharply reduces Neg-Acc (78.35 to 41.60) while localisation and Recall remain similar, demonstrating explicit encoding absence in the referring score without harming ranking. Finally, replacing InterpIoU coupled with the matching strategy, with the original GIoU implementation in GroundingDINO \cite{liu2024grounding} drops mIoU by 9.53 and Top-1 consequently, particularly for smaller instances.
\begin{table} 
    \Huge
    \centering
    \setlength{\tabcolsep}{3pt} 
    \renewcommand{\arraystretch}{1.1} 
    \caption{Ablation study of the proposed framework. We evaluate the impact of different text processing strategies (Sentence-only vs. Word-only), removal of the Query Projection, removal of the Hierarchical Constraint, and the impact of the InterpIoU coupled with the matching strategy. All metrics are reported on the Test set.}
    \label{tab:ablation_hrs_clean}
    \resizebox{0.6\linewidth}{!}{
    \begin{tabular}{lccccc}
    \toprule
    \textbf{Method} & \textbf{Top-1} & \textbf{Top-5} & \textbf{R@0.5} & \textbf{mIoU} & \textbf{Neg-Acc} \\
    \midrule
    Sentence-only               & 49.29 & 74.06 & 50.53 & 49.35 & 74.12 \\
    Word-only                   & 36.19 & 67.61 & 44.20 & 39.22 & 71.55 \\
    w/o Query Projection        & 33.20 & 58.02 & 38.43 & 35.39 & 69.83 \\
    w/o Hierarchical Constraint & 59.87 & 80.57 & 53.87 & 55.83 & 41.60 \\
    w/o InterpIoU               & 49.15 & 73.28 & 44.61 & 47.72 & 75.88 \\
    \hline
    \rowcolor{gray!10}
    \textbf{Full Model}         & \textbf{62.42} & \textbf{82.45} & \textbf{55.44} & \textbf{57.25} & \textbf{78.35} \\
    \specialrule{.1em}{.05em}{.05em}
    \end{tabular}
    }
\end{table}

\vspace{-8mm}
\section{Conclusion, Limitations and Future Work} \vspace{-6mm}
We introduce gRef-CW, a real-world Crop and Weed dataset with 78k target instances and 82k referring expressions, including negative sentences, capturing extreme scale variation, dense scenes, and fine-grained crop–weed ambiguity. Evaluations of existing VG methods reveal substantial gaps in handling these challenges. To address this, we propose Weed-VG, a modular framework that is designed to integrate existing grounding models. Within Weed-VG, we introduce HRS for target existence and instance ranking, along with interpolation-driven regression to manage scale variation and high scene density. Experiments on standard grounding and negative-accuracy metrics demonstrate that explicitly modelling existence alongside instance ranking leads to more robust and reliable performance in crop and weed scenarios. \\
\textbf{Limitations and Future Work.} This work has several limitations. Our findings are constrained by the image sources, which are limited to top-down viewpoints, and by the quality of initial proposals. Template-based annotations restrict the evaluation of free-form expressions and negations, and performance still drops in extremely dense scenes (>30 instances). While gRef-CW provides a starting point, larger-scale, agri-specific VG datasets are required. More broadly, applying the modular framework of Weed-VG to standard grounding benchmarks is possible, but the multi-level labelling scheme, capturing both target non-existence and instance-level referrals, requires significant annotation effort. Future work includes exploring efficient annotation strategies and validating the HRS approach on broader gVG benchmarks.



%
%
\bibliographystyle{splncs04}
\bibliography{main}

\clearpage
\appendix

\end{document}